\documentclass{article} %
\usepackage[preprint]{colm2025_conference}
\usepackage{comment}
\usepackage{graphicx}
\usepackage{microtype}
\usepackage{hyperref}
\usepackage{amsmath}
\usepackage{cleveref}
\usepackage{url}
\usepackage{booktabs}

\usepackage[edges]{forest}
\usepackage{lineno}

\author{
Zhouhang Xie$^1$, Junda Wu$^1$, Yiran Shen$^1$, Yu Xia$^1$, Xintong Li$^1$, Aaron Chang$^2$,\\
\textbf{Ryan Rossi$^3$, Sachin Kumar$^4$, Bodhisattwa Prasad Majumder$^5$, Jingbo Shang$^1$,}\\
\textbf{Prithviraj Ammanabrolu$^1$, Julian McAuley$^1$}\\[1ex]
$^1$University of California, San Diego \quad $^2$University of California, Los Angeles \quad\\
$^3$Adobe Research \quad $^4$The Ohio State University \quad $^5$Allen Institute for AI\\
\texttt{\{zhx022,rsurana,jes038,yux078,prithvi,jmcauley\}@ucsd.edu} \\
\texttt{aaronchang21@g.ucla.edu}, \texttt{\{ryrossi\}@adobe.com}\\
\texttt{kumar.1145@osu.edu, bodhisattwam@allenai.org}
}

\definecolor{darkblue}{rgb}{0, 0, 0.5}
\hypersetup{colorlinks=true, citecolor=darkblue, linkcolor=darkblue, urlcolor=darkblue}
\definecolor{myblue}{RGB}{112, 156, 156}
\definecolor{lightblue}{RGB}{237, 250, 252}

\title{A Survey on Personalized and Pluralistic Preference Alignment in Large Language Models}

\begin{document}

\ifcolmsubmission
\linenumbers
\fi

\maketitle

\begin{abstract}

Personalized preference alignment for large language models (LLMs), the process of tailoring LLMs to individual users' preferences, is an emerging research direction spanning the area of NLP and personalization. 
In this survey, we present an analysis of works on personalized alignment and modeling for LLMs. 
We introduce a taxonomy of preference alignment techniques, including training time, inference time, and additionally, user-modeling based methods. 
We provide analysis and discussion on the strengths and limitations of each group of techniques and then cover evaluation, benchmarks, as well as open problems in the field.

\end{abstract}

\section{Introduction}

Recently, significant progress has been made in aligning LLMs to the \textit{overall} preferences of users~\citep{zhao2024surveylargelanguagemodels, shen2023largelanguagemodelalignment}.
However, prior works show that there is no one-size-fits-all solution for preference alignment~\citep{sorensen2024roadmappluralisticalignment, kirk2024benefits}.
Intuitively, while there are universal preferences that are shared across users, such as ``it is good to respond in the tone of a friendly and helpful assistant'', user preferences are often also individualized and use-case dependent (i.e., contextual).
For example, users might have different preferences towards the tone and style of responses~\citep{jang2023personalized}, and even for the same user, the preferred style of response would change depending on the context of interaction, akin to prior works in contextualized personalization~\citep{meng2022contextualizedRecsys}.
For example, an expert might have different preferences from a beginner when asking the same question about a concept, and even the same person might have different preferences depending on when they interact with the system.

Along with the emergence of these challenges, the notion of personalization in the context of NLP and LLM research has recently attracted increasing attention within the NLP and machine learning communities~\citep{sorensen2024roadmappluralisticalignment, kirk2024benefits, flek-2020-returning}.
At first glance, personalization seems to be an intuitive solution for catering LLMs to individualized and contextual preferences described above.
However, in practice, the notion of personalization for LLMs is convoluted, ranging from applications of LLMs to classical personalization tasks~\citep{tan2023user, chen2024large} to role-playing and simulation of individual human behaviors~\citep{chen2024oscarsaitheatersurvey, mou2024individualsocietysurveysocial}.
As we shall discuss later (\Cref{sec:user_model_per_align_discussion}), not all types of personalization benefit individual users in a conversation setting.

\begin{figure}[tb]
\centering
\includegraphics[scale=0.6]{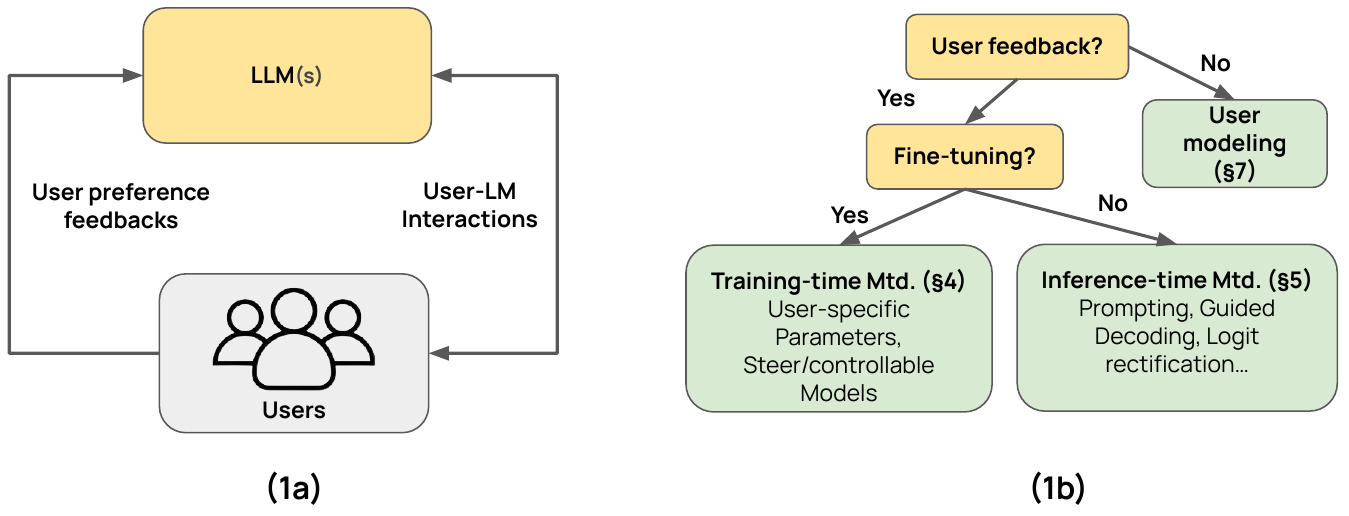}
\caption{{
(1a) Overview on personalized preference alignment for LLMs. This includes training (\Cref{sec:training_time_alignment}) and test-time (\Cref{sec:test_time_alignment}) methods, leveraging various feedbacks such as verbal feedback and choices. 
(1b) An over-simplified decision tree for determining the class of method to use for personalized preference alignment.}
}
\label{fig:figure_1}
\end{figure}

In this work, we focus on personalized and pluralistic preference alignment, a specific notion of personalization that aims at adapting an LLM's behavior to dynamic user \textit{preferences} across individuals, groups, and contexts to enhance user satisfaction.
We start by defining the problem formulation for personalized preference alignment (\Cref{sec: problem_formulation}), then introduce an intuitive technique taxonomy covering training and test-time methods for preference alignment (\Cref{sec:training_time_alignment},~\Cref{sec:test_time_alignment}).
We then discuss the connection between user modeling and personalized preference alignment (\Cref{sec:user_model_per_align_discussion}), and discuss benchmark and evaluation (\Cref{sec:benchmarks_and_evaluation}) as well as open problems in the field (\Cref{sec:future_works_emerging_directions}).

\textbf{What's covered?} Broadly speaking, as shown in~\Cref{fig:figure_1}, personalization in the context of preference alignment can be categorized into two classes of methods: Training-time (\Cref{sec:training_time_alignment}) and inference-time (\Cref{sec:test_time_alignment}) personalization that leverages personalized user feedback at different stages of LLMs' life cycle. 
Additionally, when user preference feedback is absent, personalized adaption of LLMs and LLM-based systems can sometimes also be achieved by modeling the user, which we discuss in~\Cref{sec:user_model_per_align_discussion}.

\textbf{What's not covered.}
Under the theme of LLMs and personalization, there are other relevant areas such as user-behavior modeling (i.e., systems that predict the behavior of users)~\citep{tan2023user, wu2024surveylargelanguagemodels, chen2024large}, user-group-behavior modeling (i.e., LLM role-playing)~\citep{chen2024oscarsaitheatersurvey, chen2024oscarsaitheatersurvey, tseng2024two}, and LLMs \textit{for} personalized recommender systems (e.g., LLMs as a component in product recommender systems)~\citep{tan2023user, wu2024surveylargelanguagemodels, chen2024large}.
However, these lines of work does not involve directly catering the behavior of LLMs or LLM-based conversational systems to user preferences.
We point interested readers to related surveys (\Cref{sec:related_surveys_key_differences}) on these topics.

\textbf{Contribution statement.} Despite numerous recent efforts on consolidating works that involve LLM and personalization~\citep{sorensen2024roadmappluralisticalignment, kirk2024benefits, zhang2024personalizationlargelanguagemodels, wu2024personalizedmultimodallargelanguage, tan2023user, wu2024surveylargelanguagemodels, chen2024large}, we show personalized alignment is an emerging valuable research direction, with its own methods and evaluation paradigm.
To this end, this survey provides a comprehensive overview of personalized and pluralist alignment while establishing its differences and connections to adjacent domains.
Further, personalized preference alignment is an emerging domain with no universally acknowledged evaluation and benchmarks.
To make it easier for future research to access existing evaluation schemes, we provide an overview of the current state of evaluation methods.
Overall, this survey serves as an up-to-date resource for practitioners and researchers working on personalized and pluralistic alignment and, more broadly, LLM personalization and alignment.

\section{Related Surveys and Key Differences}
\label{sec:related_surveys_key_differences}

\textbf{User modeling and personalizing LLMs to simulate user behavior.} 
Recently, a few position papers have discussed the concept of personalized preference
alignment~\citep{sorensen2024roadmappluralisticalignment, kirk2024benefits}. However, these works focus on discussing the implications without summarizing current progress.
There have also been surveys on personalizing LLMs~\citep{zhang2024personalizationlargelanguagemodels, wu2024personalizedmultimodallargelanguage}, but they do not differentiate between the replicating user behavior to catering for user preferences.
Finally, another stream of recent works focuses on LLM-based role-laying~\citep{tseng2024two}. 
However, their focus is still on empowering LLMs to replicate certain user groups' behavior. 
In contrast, our work focuses on summarizing the progress with respect to personalized preference alignment.

\textbf{Application of LLMs to downstream personalization tasks.} There are also a few surveys on the \textit{application} of LLMs to other tasks related to personalization, such as recommender systems and user modeling~\citep{tan2023user, wu2024surveylargelanguagemodels, chen2024large}, personalized wearable devices~\citep{li2024personal}, personalization on the web~\citep{chen2024large}, narrative-drive recommendation~\citep{mysore2023largelanguagemodelaugmented}, and LLM-based role-laying~\citep{chen2024oscarsaitheatersurvey}, which are different from the scope covered in this work since the goal of these lines of research is not to build an optimal LLM-based dialogue system.
We provide further clarification of our scope in~\Cref{sec: problem_formulation}, and discuss in more detail the relationship between (personalized) user modeling and preference alignment in~\Cref{sec:user_model_per_align_discussion}.

\section{Problem Formulation and Techniques Taxonomy}
\label{sec: problem_formulation}

\begin{figure*}[t!]
\centering
\begin{forest}
for tree={   
font=\fontsize{6}{6}\selectfont,
draw=myblue, semithick, rounded corners,
       minimum height = 1.ex,
        minimum width = 2em,
    anchor = west,
     grow = east,
forked edge,        %
    s sep = 0.5mm,    %
    l sep = 4mm,    %
 fork sep = 2mm,    %
           }
[A Survey of Personalized Alignment and Modeling, rotate=90, anchor=center
    [Test-time Personalized Alignment \\ (\S \ref{sec:test_time_alignment}), fit=band, text width=1.3cm, text centered, anchor=center
        [Logit Rectification and Re-Alignment (\Cref{sec:test_time_alignment}.3), text width=2.8cm, l sep = 2mm
            [{E.g., EFT \citep{mitchell2023emulator} , Proxy-Tuning  \citep{liu2024tuning}, MOD \citep{shi2024decoding}, DeRa \citep{liu2024decoding}, Alginer\citep{ji2024aligner}, MetaAlginer \citep{yang2024metaaligner}, Amulet \citep{zhang2025amulet}}, text width=7cm, fill=lightblue]
        ]
        [Reward and Value-Guided Decoding (\Cref{sec:test_time_alignment}.2), text width=2.8cm, l sep = 2mm
            [{E.g., ARGS \citep{khanov2024args}, RAD \citep{deng2023reward}, CARDS \citep{li2024cascade}, DeAL \citep{huang2024deal}, GenARM \citep{xu2024genarm} , PAD \citep{chen2024pad}, CD \citep{mudgal2023controlled}, PPO-MCTS \citep{liu2024don}, IVG \citep{liu2024inference}, VAS \citep{han2024value}}, text width=7cm, fill=lightblue]
        ]
        [Prompting-based \\(\Cref{sec:test_time_alignment}.1), text width=2.8cm, l sep = 2mm
            [{E.g., URIAL \citep{lin2023unlocking}, RAIN \citep{li2023rain}, OPO \citep{xu2023align}, LaMP \citep{salemi2023lamp}, PEARL \citep{mysore2023pearl}, FERMI \citep{kim2024few}, BPO \citep{cheng2023black}, Rewrite Prompt \citep{li2024learning}}, text width=7cm, fill=lightblue]
        ]
    ]
    [Training-time Personalized Alignment \\ (\S \ref{sec:training_time_alignment}), fit=band, text width=1.3cm, text centered, anchor=center
        [Steer/controllable Models \\(\Cref{sec:training_time_alignment}.2), text width=2.8cm, l sep=2mm
            [{E.g., LLMs and reward models that are adaptable conditioned on user stated system prompts~\cite{lee2024aligning}, user persona~\cite{wang2024learningpersonalizedalignmentevaluating}, user group identifiers~\citep{kumar2024compo}, as well as infered user-specific latent~\citep{poddar2024personalizing, chen2024pal} or explicit~\citep{balepur2025whose} personas.}, text width=7cm, fill=lightblue]
        ]
        [User-spefic Parameters \\(\Cref{sec:training_time_alignment}.1), text width=2.8cm, l sep=2mm
            [{E.g., User-specific embeddings~\citep{li2024personalized}, steering vectors~\citep{cao2024personalized}, PEFT modules~\citep{Tan2024DemocratizingLL, tan2024personalized}, soft prompts~\citep{huang2024selective}, inference heads~\citep{Zhuang2024HYDRAMF}, and mixture of preference experts~\citep{zhou2024orchestrating}.}, text width=7cm, fill=lightblue]
        ]
    ]
]
\end{forest}
\caption{Technique taxonomy on personalized and pluralistic preference alignment.}
\label{fig:taxonomy}
\end{figure*}
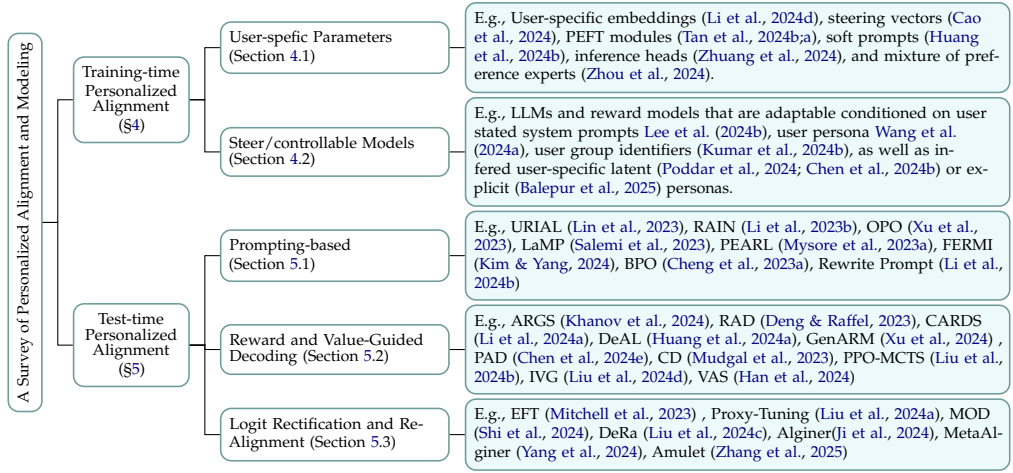

\subsection{Problem Statement}
We begin by introducing the notion of personalization in the context of preference alignment, as illustrated in~\Cref{fig:figure_1}-a. In this setting, system deployers aim to build an LLM (or LLM-based system) that caters specifically to each individual user's unique preferences.

For clarity, let \(\mathcal{X}\) denote the input space (e.g., user queries or contextual prompts), and \(\mathcal{Y}\) denote the output space (e.g., responses generated by the LLM).
We formalize the personalized reward function as
\[
r: \mathcal{X} \times \mathcal{Y} \times \mathcal{U} \rightarrow \mathbb{R},
\]
which measures how well an LLM's response \(y\) to an input \(x\) satisfies the unique preferences of user \(u\).
Our goal is to learn a set of individualized policies \(\{\pi_u\}_{u \in \mathcal{U}}\), where each policy \(\pi_u\) is tailored exclusively to user \(u\).
Such an objective can be expressed as:
\begin{equation}
     \pi_u^* = \arg\max_{\pi_u} \; \mathbb{E}_{x \sim \mathcal{X},\, y \sim \pi_u(\cdot \mid x,u)} \left[ r(x,y,u) \right] \quad \forall u \in \mathcal{U}
\label{eq:problem_statement_individual}
\end{equation}
In other words, for each user \(u\), the optimal policy \(\pi_u^*\) is one that, in expectation over the distribution of inputs, generates responses that best satisfy that user's individual preferences. In practice, this set of LLM-based policies can often have shared parameters, or are built from a single model conditioned on different user information in the prompt, as we shall later discuss in~\Cref{sec:training_time_alignment} and~\Cref{sec:test_time_alignment}.

\subsection{Technique Overview}
\label{sec:technique_taxonomy}

The problem formulation discussed above yields a straight-forward way to partition existing methods, as shown in~\Cref{fig:taxonomy}.
Specifically, in order to obtain a LLM-based policy that cater to each individual's preferences, it is crucial to be able to adapt the LLM itself depending on the interacting user.
Naturally, this can be achieved both by \textit{training} models with user-specific parameters (\Cref{sec:training_time_alignment}.1) or making the model steerable with respect to user inputs (\Cref{sec:training_time_alignment}.2), as shown in~\Cref{fig:figure_1}-b.
On the other hand, fine-tuning or adapting LLMs is frequently costly.
This challenge motivates another category of works that aims at influencing a pre-trained LLMs' behavior at inference time, with methods such as prompting (\Cref{sec:test_time_alignment}.1), controlled-decoding (\Cref{sec:test_time_alignment}.2), and logit manipulation (\Cref{sec:test_time_alignment}.3).
Additionally, we note that there are personalization techniques that nevertheless improve base LLMs towards the goal of catering to individual preferences without using user feedback, such as building user-specific memories (e.g., \citet{yuan-etal-2025-personalized, zhang-etal-2022-history}) following the assumption that users generally likes to be remembered.
We provide discussion of this complementary class of valuable techniques in~\Cref{sec:user_model_per_align_discussion}.

\subsection{Granularity of Personalization}

While personalized preference alignment aims to cater to, ultimately, individualized preferences, personalization typically suffers from the issue of sparse feedback~\citep{li2023recentdevelopmentsrecommendersystems}.
Specifically, an individual user's interactions with the system are frequently too few to allow meaningful learning to happen.
To this end, similar to works in adjacent personalization areas such as recommender systems~\citep{li2023recentdevelopmentsrecommendersystems}, it is often helpful to exploit the fact that there can be \textit{groups} of users that shares similar preferences, thus bypassing the feedback sparsity issue.
Similar to these prior works, personalized LLM alignment can also happen on different granularity: individual users' levels and user groups based on user profiles and social relationships~\citep{sorensen2024roadmappluralisticalignment, kumar2024compo}.
We note that there is also a special case of ``contextual" shared preference between user groups, where a set of users momentarily shares preferences based on their purpose of interacting with LLMs.
For example, a group of users that are seeking help from an LLM-based therapy chatbot may collectively wish the LLM to act as a helpful therapist~\cite{Stade2024LargeLM}.
We provide discussions on this special case in~\Cref{sec:user_model_per_align_discussion}.

\section{Training-time Personalized Alignment}\label{sec:training_time_alignment}

To effectively adapt LLMs and LLM-based systems to personalized user preferences, a straight-forward solution is to develop specialized models via training. 
In this section, we introduce two popular classes of techniques: building models with user-specific parameters and training models that are sensitive to input user preferences.

\subsection{Learning from Feedbacks with User-specific Parameters}

\textbf{Motivation.} One of the most straightforward ways to build LLM that caters to individualized user preference is by keeping separate parameters for each user (or user group), effectively learning a set of policies whose behavior slightly deviates from each other without explicitly specified user preference under the standard RLHF setting.

\textbf{Comparative Analysis.} Addressing the challenge of inferring user-specific preferences without requiring explicit specification, \citet{li2024personalized} employs a lightweight user model to learn from human feedback, while
\citet{cao2024personalized} follows a similar setting, but learn per-user steering-vectors to achieve personalization.
Following similar intuition, \citet{Tan2024DemocratizingLL} and \citet{tan2024personalized} advocates for a dedicated per-user PEFT module that stores individual behavior patterns. 
Other than adapter modules, \citet{huang2024selective} explores a complimentary parameter-efficient personalization approach via soft prompts.
\citet{Zhuang2024HYDRAMF} introduces HYDRA, a factorization framework that couples a shared base model with user-specific heads, yielding notable improvements over prompt-based methods.
\citet{zhou2024orchestrating} introduces RLPHF, which merges outputs from specialized expert LLMs using a lightweight Preference Control Model (PCM) that dynamically adjusts token predictions based on user context; i.e., mixture of preference experts.
Finally, in contrast to learning user-specific parameters in LLMs, \citet{park2024rlhf} explored learning multiple reward models to handle the trade-off between bias and variance in personalized preference alignment.

\begin{figure}[btp]
    \centering
    \includegraphics[trim=0cm 0cm 0cm 0cm, clip, width=1.0\linewidth]{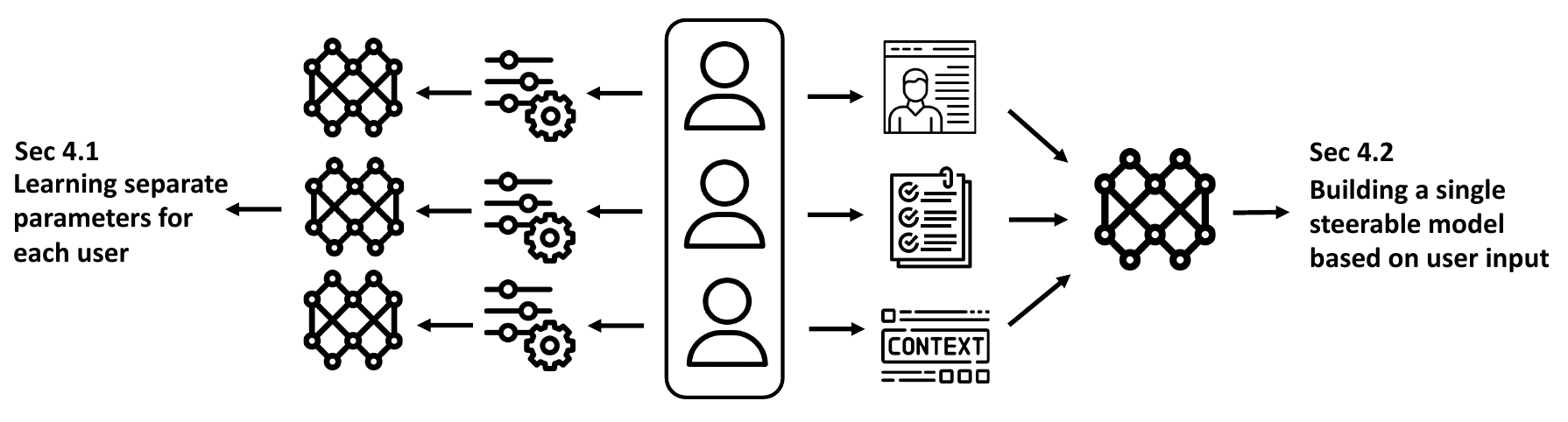}
    \caption{Personalized language model alignment during training time.}
    \label{fig:train-time}
\end{figure}

\textbf{Limitations.} Despite these advances, existing personalized RLHF frameworks often struggle to simultaneously ensure personalized adaptation and global model performance, with many approaches relying on complex multi-stage processes or additional components that may hinder model scalability \citep{park2024principled, han2024value, lee2024bapobaseanchoredpreferenceoptimization}.
Furthermore, challenges remain in robustly handling heterogeneous and strategic feedback while integrating efficient privacy-preserving techniques, pointing to the need for more streamlined and resilient solutions \citep{salemi2024comparing} such as federated personalized alignment~\citep{zhang2024personalized, jiang2024personalized, jiang2024personalizedwirelessfederatedlearning}.

\subsection{Building Steerable Model that Adapts Responses Based on User Input}

\textbf{Motivation.} Another popular choice for personalization is building a single base model that's steerable~\citep{sorensen2024roadmappluralisticalignment}, where are \textit{single} models' behavior changes based on the user it is interacting with, similar to prior research for controlled text generation~\citep{hu2017towardcontroll}.
This line of research often assumes user inputs relevant to individualized preferences are available, such as explicitly stated preferences or personas that imply user preferences are available at inference time, bypassing the need for user-specific parameters.

\textbf{Comparative Analysis.}
Following the assumptions that user can provide their own preferences as prompts to LLM, \citet{lee2024aligning} train an instruction-following LLM that can adapt to user-written values in system prompt via data synthesis,
\citet{wang2024learningpersonalizedalignmentevaluating} builds base models geared towards llm-as-a-judge~\citep{zheng2023judging} use-cases that can adapt to explicitly stated user personas.
Aside from building steerable LLM policies, such adaptability can also be built into reward models.
For example, \citet{pitis2024improving} builds context-conditioned reward models, which are then used for personalized (i.e., context-conditioned) alignment.
Similarly, \citet{kumar2024compo} builds reward models conditioned on user group identifiers on Reddit to achieve personalization.
We note that there are works that attempt to infer latent user representations when users don't explicitly state their preference, but still build steerable models based on these latent representations.
For example, \citep{balepur2025whose} infer user persona from choices using LLMs, and train model to adapt to those personas.
In contrast, \citet{chen2024pal} adopts a plurality-based approach, using ideal point and mixture modeling to learn a common latent preference space that generalizes to new users,
Finally, \citep{poddar2024personalizing} also learns user latent factors, but instead opts to use user-conditional reward models to achieve personalization.

\textbf{Limitations.} While alleviating the need for maintaining user-specific parameters, this class of methods often depends on user such as personas and verbal preferences, which are unavailable in standard RLHF settings. 
Further, collecting datasets with diverse user personas itself is poses a challenge, which we discuss extensively in~\Cref{sec:benchmarks_and_evaluation}.

\section{Test-time Personalized Alignment}\label{sec:test_time_alignment} 

The growing need for language models that can adapt on the fly to diverse user preferences—while keeping compute costs manageable and simplifying deployment—has spurred a variety of test‐time alignment techniques. These methods adjust the decoding process, without altering the underlying model parameters, to steer outputs toward desired behaviors and make debugging easier. We broadly categorize these approaches into three groups: prompting and context optimization, reward- and value-guided decoding methods, and rectification-based decoding and correction.

\subsection{Prompting-based Alignment Methods}
\begin{figure}[btp]
    \centering
    \includegraphics[trim=0cm 0cm 0cm 0cm, clip, width=0.9\linewidth]{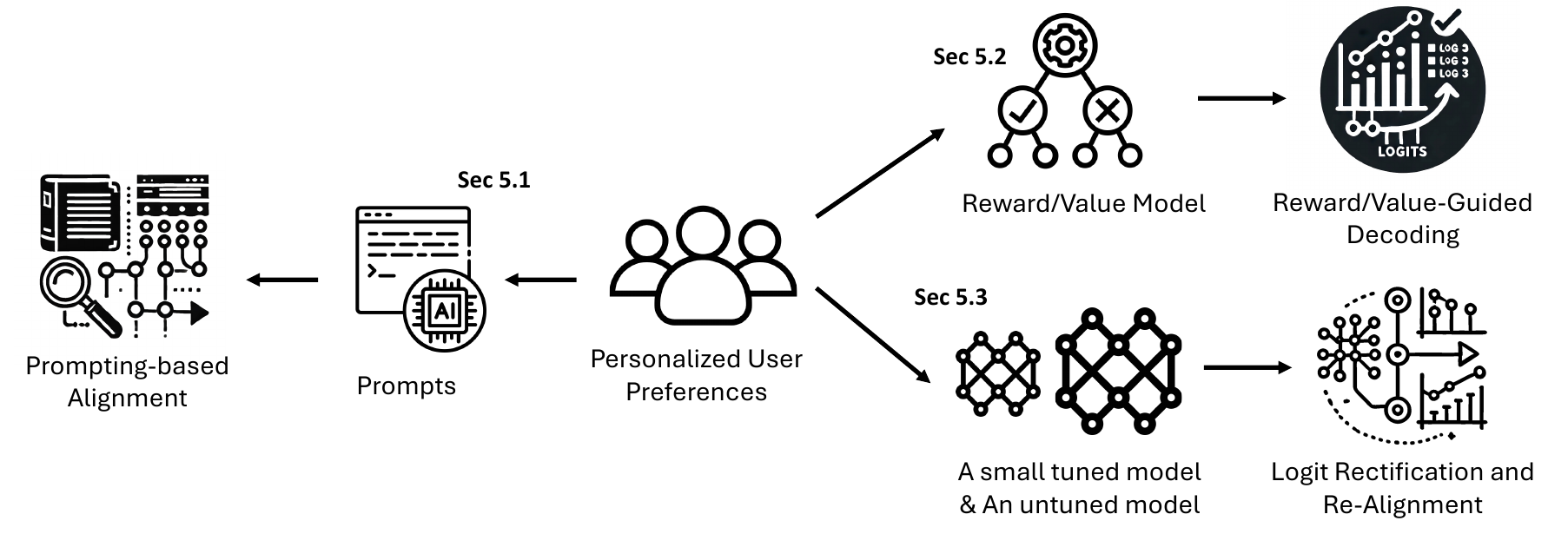}
    \caption{Personalized language model alignment during test time.} 
    \label{fig:test-time}
\end{figure}

\textbf{Motivation.}  
Prompting-based approaches modify the input context—via in-context examples, retrieval, or prompt rewriting—to elicit behavior that mirrors specific user preferences. This training-free strategy is particularly attractive for personalization, as it can quickly adapt to new or evolving user profiles (see Figure~\ref{fig:test-time}).

\textbf{Comparative Analysis.}  
A central challenge for prompting-based methods is achieving personalized alignment without the need for full-scale fine-tuning. \citet{lin2023unlocking} and \citet{li2023rain} propose methods that harness a small number of stylistic examples or self-correction mechanisms to nudge LLMs toward outputs that align with individual user styles. Similarly, \citet{xu2023align} dynamically incorporate external memories to retrieve rules tailored to diverse social norms, while \citet{salemi2023lamp} and \citet{mysore2023pearl} further personalize responses by retrieving user-specific items to augment the prompt. In parallel, prompt optimization techniques by \citet{kim2024few}, \citet{cheng2023black}, and \citet{li2024learning} iteratively refine prompts based on misaligned outputs, thereby progressively incorporating user feedback. 

\textbf{Limitations.}  
Despite their promise for personalization, these methods are limited by the fixed context length and increased computational overhead when handling complex, dynamic user profiles. Additionally, the reliance on relatively static representations of user preferences can hinder real-time adaptation to rapidly changing individual needs.

\subsection{Reward and Value-Guided Decoding Methods}
\textbf{Motivation.}  
Reward and value-guided decoding methods integrate personalized alignment objectives directly into the LLM decoding by adjusting token probabilities based on reward signals or value functions. This approach supports fine-grained, token-level personalization and enables LLMs to adapt their outputs on the fly to meet individual user preferences.

\textbf{Comparative Analysis.}  
One of the major challenges is balancing high-reward personalized output with maintaining natural language fluency. \citet{khanov2024args} and \citet{deng2023reward} propose frameworks that incorporate reward signals into the decoding process, which can be tuned to reflect personalized quality metrics. Building on this, \citet{li2024cascade} and \citet{huang2024deal} introduce strategies that break the generation process into segments, allowing for iterative refinement toward personalized and fluent outputs. Complementing these reward-guided methods, \citet{xu2024genarm} and \citet{chen2024pad} focus on scalable, test-time alignment by employing autoregressive reward models and personalized reward modeling, respectively. In parallel, value-guided techniques offer another path to personalization: \citet{mudgal2023controlled} and \citet{liu2024don} integrate auxiliary value functions during decoding, whereas \citet{liu2024inference} and \citet{han2024value} demonstrate that combining implicit and explicit value guidance can further enhance the model’s responsiveness with respect to users' personalized preferences and values.

\textbf{Limitations.}  
A key limitation of reward and value-guided methods is their reliance on pre-trained reward or value models, which introduces challenges such as limited transparency and sensitivity to adversarial inputs. Additionally, the need for real-time evaluation of reward signals and value functions often results in significant computational overhead, making it challenging to scale these methods for high-throughput, low-latency applications.

\subsection{Logit Rectification and Re-Alignment Approaches}
\textbf{Motivation.}  
Logit rectification approaches modify the internal decision process of LLMs by integrating corrective signals—often from smaller, fine-tuned models or auxiliary modules—directly into the decoding stage. This strategy enables personalized re-alignment without retraining the entire model, thereby offering a lightweight path to tailor outputs to individual user needs.

\textbf{Comparative Analysis.}  
A key challenge in this category is to harness the complementary strengths of large pretrained models and smaller, aligned models. \citet{mitchell2023emulator} introduce emulated fine-tuning, which decouples the contributions of pre-training and fine-tuning by combining the knowledge of a large model with the behavioral adjustments of a small model; a special case, LM up-scaling, demonstrates that ensembling can emulate the effects of full fine-tuning without additional training. In a similar vein, \citet{liu2024tuning} propose proxy-tuning, a lightweight algorithm that steers the output distribution of a large black-box model using the difference between the predictions of a tuned small model and its untuned counterpart. Addressing the need for multi-objective personalization, \citet{shi2024decoding} develop a method that computes a linear combination of predictions from several base models, enabling flexible adjustment of competing objectives, while \citet{liu2024decoding} propose decoding-time re-alignment to dynamically control the degree of alignment without retraining. Complementing these strategies, \citet{ji2024aligner} offers a model-agnostic correction approach that learns residuals to refine outputs, and \citet{yang2024metaaligner} extends this idea with a generalizable framework that supports multi-objective and personalized alignment by adjusting target objectives via prompt updates. Finally, \citet{zhang2025amulet} frame re-alignment as an online per-token optimization problem with a closed-form solution, achieving real-time adaptation to diverse and evolving user preferences.

\textbf{Limitations.}  
Rectification-based approaches are highly dependent on the quality and consistency of the auxiliary correction signals. Their effectiveness can be limited if the small, tuned models do not capture the full complexity of user-specific nuances, which may lead to a mismatch between the corrective adjustments and the large model's inherent output tendencies. Additionally, ensuring robust performance across a wide range of personalization scenarios without introducing extra latency remains a practical challenge.

\section{Dataset \& Evaluation}\label{sec:benchmarks_and_evaluation}

\paragraph{Datasets.}
Early attempts on constructing datasets for personalized preference alignment typically rely on data synthesization, with some recent attempts in collecting real-world data for preference alignment.
We provide an overview of relevant benchmarks and evaluations in~\Cref{tab:datasets}.
For example, \citet{cheng2023everyone} introduce the Domain-Specific Preference (DSP) dataset by augmenting the Alpaca instruction corpus~\citep{alpaca} with answers targeting different domain (e.g., Academia or Entertainment).
\citet{jang2023personalized} introduce Personalized Soups (P-SOUP), where Alpaca instruction prompts~\citep{alpaca} are answered by an LLM in contrasting styles along three dimensions (expertise level, verbosity, and tone) and ranked based on predefined style preferences.
On a much larger scale, \citet{lee2024aligning} constructed the Multifaceted Collection dataset, containing about 65k instructions, each paired with three LLM-generated responses that vary along thousands of dimensions.
Finally, \citet{zollo2024personalllm} present PersonalLLM, an open benchmark that generate preference rankings from a set of simulated synthetic ``users''.
In parallel, other recent works have assembled personalized human preference datasets to study personalization with real users . The PRISM Alignment dataset \citep{kirk2024prism} is a notable example, recording 8,011 live chat interactions between 1,500 users across 75 countries, with each user’s persona profile and preference feedbacks (ratings), providing a complementary setting for evaluating personalized preference alignment.

\begin{table*}[!t]
\centering
\resizebox{\textwidth}{!}{%
\begin{tabular}{l|l|l|l}
\toprule
\textbf{Reference} & \textbf{Task/Data} & \textbf{Metric (Dimension)} & \textbf{Notes} \\
\midrule
DSP \citep{cheng2023everyone} 
    & Prompt selection 
    & PPA (Domain: Acad., Bus., Ent., Lit. \& Art) 
    & Tailored prompts \\

P-SOUP \citep{jang2023personalized} 
    & Pairwise feedback
    & PWR (Expertise, Info., Friendliness)
    & GPT-4 simulated \\

MULTIFACETED \citep{lee2024aligning} 
    & Preference alignment
    & AAS (Human pref.)
    & Diverse preferences \\

HH-RLHF \citep{bai2022training} 
    & RLHF dataset 
    & CRS (Helpfulness, Harmlessness) 
    & Personalized dimensions \\

HelpSteer2 \citep{wang2024helpsteer2} 
    & RLHF feedback 
    & MFS (Helpfulness, Harmlessness, Humor) 
    & Humor-enhanced \\

PRISM \citep{kirk2024prism} 
    & Demographic alignment 
    & AFI (Fairness across demographics)
    & Participatory alignment \\

LaMP \citep{salemi-etal-2024-lamp} 
    & Personalized generation 
    & POQ (User-profile retrieval)
    & Retrieval-augmented \\

LongLaMP \citep{kumar2024longlamp} 
    & Personalized long-text generation 
    & POQ (User-profile retrieval)
    & Retrieval-augmented \\

PersonalLLM \citep{zollo2024personalllm} 
    & Personalization benchmark
    & PAS (Individual prefs.)
    & Beyond uniform alignment \\

ALOE \citep{wu2024aligning} 
    & Multi-turn dialogues 
    & PCS (Persona consistency)
    & Persona-specific dataset \\

PersoBench \citep{afzoon2024persobench} 
    & Persona-aware dialogue 
    & PAA (Persona awareness)
    & Zero-shot evaluation \\
\bottomrule
\end{tabular}%
}
\caption{Compressed overview of datasets and benchmarks for personalized alignment and generation.}
\label{tab:datasets}
\end{table*}

\paragraph{Evaluation in Personalized Alignment.}
Evaluating personalization in LLM alignment requires measuring how well a model’s output matches the particular preferences of a user. 
When the evaluation dataset includes response variations across known dimensions (e.g., style, tone), evaluation often uses multi-dimensional scoring or pairwise comparisons. For example, \citet{jang2023personalized} assess their proposed methods by scoring responses along each dimension separately, but assumes different users place different value on the set of dimensions.
Another popular evaluation approach is to employ LLM-as-a-judge~\cite{zheng2023judging}, but prompt the judge LLMs with pre-defined user personas \citep{wu2024aligning, lee2024aligning}.
This evaluation formulation leads to the creation of specialized evaluation models such as PerSE \citep{wang2024learningpersonalizedalignmentevaluating}, a 13B Llama-2 based evaluator fine-tuned to judge alignment with personal profiles. 

\paragraph{Limitations.}
Despite progress on model evaluations catered for personalized preference alignment, most assessments rely heavily on rule-based metrics or persona-based LLM-as-a-judge~\cite{zheng2023judging}, yielding simulations of user satisfaction. 
Since these heuristics and personas for evaluations are use-case specific, there are currently no unified evaluation across studies, posing a challenge for systematic evaluation and progress tracking. 
As the research community moves toward widely accepted multidimensional preference benchmarks and the development of publicly available evaluators, more consistent and comparable metrics are expected to emerge.

\section{On Personalized User Modeling and Personalized Preference Alignment}
\label{sec:user_model_per_align_discussion}

\textbf{Why this section?} As discussed previously in~\Cref{sec:related_surveys_key_differences}, personalized preference alignment is a subset of personalization research for LLMs.
Specifically, there is another emerging research area that focuses on LLM-based user modeling~\cite{tan2023user}, building LLM-based \textit{simulations} for individual users, which most commonly manifest as directly predicting users' responses.
While prior surveys do not differentiate between personalized user modeling and preference alignment~\citep{zhang2024personalizationlargelanguagemodels, wu2024personalizedmultimodallargelanguage, tseng2024two}, we note that these are two distinct and complementary research directions.
To this end, this section serves two purposes. 
First, we clearly distinguish between the two research directions currently studied under LLMs and personalization, making it easier for future researchers to sift through relevant literature.
Second, we discuss various ways user modeling can enable better-personalized preference alignment.

\textbf{Personalized preference alignment research benefits from personalized user modeling.}
Due to the difficulty in collecting large-scale feedbacks from real users, current research for personalized preference alignment relies heavily on user simulation.
For example, various recently proposed datasets rely on persona-grounded simulation with LLMs to build benchmarks~\citep{cheng2023everyone, jang2023personalized, lee2024aligning, zollo2024personalllm} or evaluations~\citep{zheng2023judging, wu2024aligning, lee2024aligning, wang2024learningpersonalizedalignmentevaluating} for personalized preference methods.
To this end, better simulation of diverse, realistic user behaviors naturally improves the development of better personalized alignment methods.

\textbf{Modeling individual user (sometimes) enables preference-free personalized alignment.}
Meanwhile, by modeling individual users, system deployers can still increase user satisfaction, even without directly modeling user preferences. 
For example, in areas such as personalized review prediction~\citep{xie23factual, ni-etal-2019-justifying} and recommender system~\citep{wu2024surveylargelanguagemodels}, user behavior happens to strongly correlate with user preferences, and thus better prediction of user behavior directly improves preference alignment.
Similarly, related research directions such as LLM-based chit-chat dialogue systems with personalized user memory~(e.g., \citet{yuan-etal-2025-personalized}) also better caters to the preference of individual users by remembering personal facts, even when there are no explicit modeling of user preferences.
Finally, simulating specific desired personas such as teachers~\citep{wang2024largelanguagemodelseducation}, therapists~\citep{Stade2024LargeLM}, and travel-planners~\citep{chen2024travelagentaiassistantpersonalized} also enables LLMs to better cater to the corresponding user groups, such as students, patients, and travelers.

\section{Future Works and Emerging Directions}
\label{sec:future_works_emerging_directions}

\textbf{Online and Continuous Personalized Alignment}
While existing work on personalized preference alignment primarily explores the setting of learning user preference from offline data or given explicitly stated user preference, another complementary setting is personalized LLM alignment in an online setting~\citep{chen2024onlinepersonalizingwhiteboxllms}.
Additionally, given prior success in adjacent research (such as user modeling) on continuous personalization over multiple dialogue sessions~\citep{li2024hello, zhang2023llm, zhong2024memorybank, qian2024memorag}, personalized preference alignment in a multi-session dialogue setting is a natural extension, which typically models turn-wise or user-provided preference feedback.

\textbf{Addressing long and complex user-generated value statements}
As discussed in prior sections, personalized preference alignment in LLMs frequently relies on instruction following ability of LLMs as building blocks for alignment methods, both at training time and inference time.
However, recent works show long and complex instruction-following is still an open challenge~\citep{wu2024lifbenchevaluatinginstructionfollowing, gavin2024longinschallenginglongcontextinstructionbased}.
Given the prevalence of personalized alignment methods that rely on explicit verbal preference statements (\Cref{sec:training_time_alignment}.2 and~\Cref{sec:test_time_alignment}.1), it is still unclear whether existing methods can support complex and long user value statements.
To this end, developing benchmarks and methods to further research the instruction-following ability of LLMs on complex user preference value statements can help LLM-based dialogue systems better handle rich, multifaceted user preferences.

\section{Conclusions}

In this survey, we perform a comprehensive analysis of existing methods, datasets, and benchmarks for personalized preference alignment in LLMs and LLM-based dialogue systems. 
We discuss various classes of methods and their advantages and drawbacks, covering both training and inference-time, as well as user-knowledge-based personalized preference alignment methods. 
We also discuss limitations and future directions for LLMs to cater to diverse and individualistic preferences.

\bibliography{latex/custom}
\bibliographystyle{colm2025_conference}

\appendix

\end{document}